\documentclass[runningheads]{llncs}
\usepackage{graphicx}
\usepackage{cite}
\usepackage{amsmath,amssymb,amsfonts}
\usepackage{algorithmic}
\usepackage{graphicx}
\usepackage{textcomp}
\usepackage{xcolor}
\usepackage{tabularx}

\usepackage{pifont}
\newcommand{\cmark}{\textcolor{green!80!black}{\ding{51}}}
\newcommand{\xmark}{\textcolor{red}{\ding{55}}}
\begin{document}
\title{Engagement Detection with Multi-Task Training in E-Learning Environments%\thanks{Supported by organization x.}
}
\titlerunning{ED-MTT in E-Learning Environments}

\author{Onur Copur\inst{1} \and
Mert Nak\i p\inst{2} \and
Simone Scardapane\inst{1} \and
Jürgen Slowack\inst{3}
}
\authorrunning{O. Copur et al.}
\institute{Sapienza University of Rome, Roma,  00185, Italy \\ \email{onurcopur12@gmail.com, simone.scardapane@uniroma1.it} \and
Institute of Theoretical and  
		Applied Informatics, Polish Academy of Sciences, Gliwice, 44--100, Poland \\
		\email{mnakip@iitis.pl} 
\and
Barco NV, Kortrijk, 8500, Belgium \\ \email{jurgen.slowack@barco.com}}

\maketitle              % typeset the header of the contribution
\begin{abstract}

Recognition of user interaction, in particular engagement detection, became highly crucial for online working and learning environments, especially during the COVID-19 outbreak. Such recognition and detection systems significantly improve the user experience and efficiency by providing valuable feedback. In this paper, we propose a novel Engagement Detection with Multi-Task Training (ED-MTT) system which minimizes mean squared error and triplet loss together to determine the engagement level of students in an e-learning environment. The performance of this system is evaluated and compared against the state-of-the-art on a publicly available dataset as well as videos collected from real-life scenarios. The results show that ED-MTT achieves $6\%$ lower MSE than the best state-of-the-art performance with highly acceptable training time and lightweight feature extraction. 

\end{abstract}
\section{Introduction}\label{sec:introduction}

During the COVID-19 outbreak, nearly all of the learning activities, as other meeting activities, transferred to online environments \cite{zhu2020multi}. Online learners participate in various educational activities including reading, writing, watching video tutorials, online exams, and online meetings. During the participation in these educational activities, participants show various engagement levels, e.g. boredom, confusion, and frustration \cite{daisee}. To provide feedback to both instructors and students, online educators need to detect their online learners’ engagement status precisely and efficiently. For example, the teacher can adapt and make lessons more interesting by increasing interaction, such as asking questions to involve non-interacting students. Since, in e-learning environments, students are not speaking most of the time, the engagement detection systems should extract valuable information from only visual input \cite{bestemotiw}. This makes the problem non-trivial and subjective because annotators can perceive different engagement levels from the same input video. The reliability of the dataset labels is a big concern in this setting but often is ignored by the current methods \cite{bestemotiw,wu2019multi,zhu2020multi}. Because of this, deep learning models overfit to the uncertain samples and perform poorly on validation and test sets. 

In this paper, we propose a system called Engagement Detection with Multi-Task Training (ED-MTT) \footnote{Code and pretrained model are available at https://github.com/CopurOnur/ED-MTT} 
to detect the engagement level of the participants in an e-learning environment. The proposed system first extracts features with \textit{OpenFace} \cite{OpenFace}, then aggregates frames in a window for calculating feature statistics as additional features. Finally, it uses  Bidirectional Long Short-Term memory (Bi-LSTM) \cite{LSTM} unit for generating vector embeddings from input sequences. In this system, we introduce a triplet loss as an auxiliary task and design the system as a multi-task training framework by taking inspiration from \cite{roy2021self}, where self-supervised contrastive learning of multi-view facial expressions was introduced. The reason for the triplet loss usage is based on the ability to utilize more elements for training via the combination of original samples. In this way, it avoids overfitting and makes the feature representation more discriminative \cite{Dong_2018_ECCV}. To the best of our knowledge, this is a novel approach in the context of engagement detection. The key novelty of this work is the multi-task training framework using triplet loss together with Mean Squared Error (MSE). The main advantages of this approach are as follows:
\begin{itemize}
    \item Multi-task training with triplet and MSE losses introduces an additional regularization and reduces possibly over-fitting due to very small sample size.
    \item Using triplet loss mitigates the label reliability problem since it measures relative similarity between samples.
    \item A system with lightweight feature extraction is efficient and highly suitable for real-life applications.
\end{itemize}

Furthermore, we evaluate the performance of ED-MTT on a publicly available ``Engagement in The Wild'' dataset \cite{dhall2020emotiw}, which is comprised of separated training and validation sets. In our experimental work, we first analyze the importance of feature sets to select the best set of features for the resulting trained ED-MTT system. Then, we compare the performance of ED-MTT with 9 different works \cite{wang2019bootstrap,thong2019engagement,yang2018deep,niu2018automatic, thomas2018predicting, abedi2021affect, 10.1007/978-3-030-82469-3_4, zhu2020multi, chang2018ensemble} from the state-of-the-art which will be reviewed in the next section. Our results show that ED-MTT outperforms these state-of-the-art methods with at least $6\%$ improvement on MSE. 

The rest of this paper is organized as follows: Section~\ref{sec:related} reviews the related works in the literature. Section~\ref{sec:model} explains the architectural design of ED-MTT. Section~\ref{sec:results} presents experimental results for the performance evaluation of ED-MTT and comparison with the state-of-the-art methods. Section~\ref{sec:conclusion} conclude our work and experimental results. 

\section{Related Works}\label{sec:related}

One of the first attempts to investigate the relationships between facial features, conversational cues, and emotional expressions with engagement detection is presented by D'Mello et al. in \cite{d2009multimethod}. The authors in \cite{6786307,grafsgaard2013automatically} used the Facial Action Coding System (FACS) which is a measure of discrete emotions with facial muscle movements, and point out the relation between specific engagement labels and facial actions. In Reference \cite{6786307}, Whitehill et al. showed that automated engagement detectors perform with comparable accuracy to humans. %All these works were using classical machine learning classifiers such as Gentleboost and SVM.
In \cite{8273641}, Booth et al. compared the performance of a Long-Short Term Memory (LSTM) based method with SVM and KNN methods with non-verbal features. In \cite{8560296}, Dewan et al. proposed a Local Directional Pattern (LDP) to extract person-independent edge features which are fed to a Deep Belief Network. Huang et al. \cite{8784559} proposed a model, called Deep Engagement Recognition Network (DERN), which combines temporal convolution, bidirectional LSTM, and an attention mechanism to identify the degree of engagement based on the features captured by OpenFace \cite{OpenFace}. Moreover in \cite{liao2021deep}, Liao et al. proposed Deep Facial Spatiotemporal Network (DFSTN) which is is developed based on extracting facial spatial features and global attention for sequence modeling with LSTM. Finally, in \cite{thiruthuvanathanengagement,murshed2019engagement,raorecognition}, authors used models which are based on Convolutional Neural Networks (CNN) and Residual Networks (ResNet) \cite{resnet50}. All the works above considered the engagement detection problem as a multi-class classification problem. In contrast, in this paper, we follow a more recent line of research that considers engagement detection as a regression problem, where MSE loss is used to measure a continuous distance between predicted and ground truth engagement levels. 

%A trend of recent research also considers engagement detection as a regression problem and uses MSE loss. %MSE is more suitable than categorical cross-entropy loss to engagement detection tasks since it considers the distance between engagement labels. 
%In this recent line of research, 
Yang et al. \cite{yang2018deep} also used MSE loss and developed a method that ensembles four separate LSTMs using facial features extracted from four different sources. In \cite{niu2018automatic}, Niu et al. combined the outputs of three Gated Recurrent Units (GRU) based on a 117-dimensional feature vector composed of eye gaze action units and head pose features. In \cite{thomas2018predicting}, Thomas et al. used Temporal Convolutional Network (TCN) on the same set of features as in \cite{niu2018automatic}. In previous works \cite{zhu2020multi,bestemotiw}, the most common ways to overcome over-fitting is data augmentation and cross-validation training. Some other works \cite{wang2019bootstrap,abedi2021affect} consider imbalanced sampling \cite{JMLR:v18:16-365} and using weighted/ranked loss functions. Moreover, some works also consider spatial dropout and batch normalization as a regularization technique  \cite{thomas2018predicting,chang2018ensemble}. All the previous studies focus on small sample sizes and imbalanced labels but none of them consider the reliability of the labels. On the other hand, in this paper, ED-MTT aims to handle both overfitting and label reliability at the same time via multi-task training with triplet loss. 

\section{Architectural Design for Engagement Detection with Multi-Task Training}\label{sec:model}

\begin{figure*}[h!] 
	\centering
	\includegraphics[width=\textwidth]{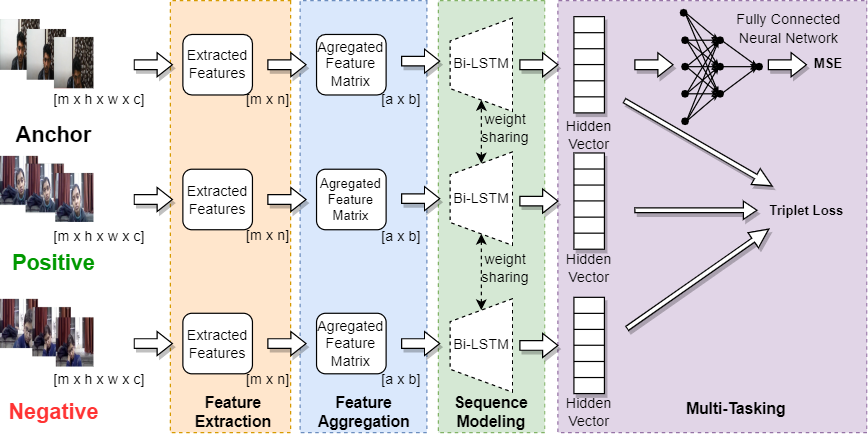}\vspace{-0.3cm}
\caption{The training architecture for ED-MTT.}
\label{fig: triplet_loss}
\end{figure*}

We now present our architectural design as well as the multi-tasking with the combination of MSE and triplet loss for training, which are the main contributions of this work. To this end, Fig.~\ref{fig: triplet_loss} displays the training architecture of ED-MTT that consists of four main parts: Feature Extraction, Frame Aggregation, Sequence Modeling, and Multi-Tasking. 
The inputs of this architecture are three batches of samples as Anchor, Positive and Negative. In each batch, each sample is the sequence of images which is obtained by segmenting a video into $m$ frames each of size $h \times w \times c$, where $h$ denotes the height in pixels, $w$ denotes the width in pixels and $c$ denotes the number of color channels of each frame, where RGB color space is used. %with a window size of $m$ and with $j$ overlapping frames where $ 1 \leq m \leq l, m \in \mathbb{Z} $ and $ 0 \leq j \leq m-1, j \in \mathbb{Z} $ 
During the training with this approach, each sample $s$ in the anchor batch is assumed to have a labeled engagement level $E^s$ between $0$ and $1$. For each $s$, $E^s$ is assigned into either low engagement or high engagement classes. To this end, if $E^s<0.5$, $s$ is assigned into the low engagement class; otherwise, i.e. $E^s\geq 0.5$, $s$ is assigned into the high engagement class. Then, for each sample $s$ in the anchor batch, the positive batch contains a random sample from the same engagement class of $s$ while the negative batch contains a random sample from the opposite engagement class of $s$.

Furthermore, the outputs of the architecture in Fig.~\ref{fig: triplet_loss} are the MSE and Triplet Loss which are combined to train the Bi-LSTM model. Note that during inference, the engagement level prediction is the output of the fully connected neural network. While creating a multi-task learning problem through triplet loss, which aims to prevent overfitting due to the very few samples available for engagement detection during e-learning, we are able to perform regression for continuous engagement levels using MSE. In the rest of this section, we explain each part of the training architecture.

\subsection{Feature Extraction}

In order to narrow down the feature space by extracting the important features from the sequence of video frames, we first determine the features that are related to the engagement level of a subject. Accordingly, as done in \cite{8784559,yang2018deep,niu2018automatic,thomas2018predicting,bestemotiw}, we consider 29 features which are related to eye gaze, head pose, head rotation, and facial action units. We extract these features with OpenFace which provides many different facial features \cite{OpenFace} and can be described as  
\begin{equation}
    \mathbf{Y}^s_{m \times n} = \text{OpenFace}(\mathbf{X}^s_{mhwc}),
\end{equation}
where $\mathbf{X}^s_{mhwc}$ is the tensor of frame sequences at sample $s$, and $\mathbf{Y}^s_{m \times n}$ is the matrix of sequence of features at sample $s$, where the $(i, j)$-th element of $\mathbf{Y}^s_{m \times n}$ the feature $i$ for frame $j$.  

In the result of feature extraction, the eye gaze-related features are, \textit{gaze\_0\_x, gaze\_0\_y, gaze\_0\_z} which are eye gaze direction vectors in world coordinates for the left eye and \textit{gaze\_1\_x, gaze\_1\_y, gaze\_1\_z} for the right eye in the image. The head pose-related features are \textit{pose\_Tx, pose\_Ty, pose\_Tz} representing the location of the head with respect to the camera in millimeters (positive Z is away from the camera). \textit{pose\_Rx, pose\_Ry, pose\_Rz} indicates the rotation of the head in radians around x,y,z axes. This can be seen as pitch (Rx), yaw (Ry), and roll (Rz). The rotation is in world coordinates with the camera being the origin. Finally, the following 17 facial action unit intensities varying in the range $0-5$ are used: \textit{AU01\_r, AU02\_r, AU04\_r, AU05\_r, AU06\_r, AU07\_r, AU09\_r, AU10\_r, AU12\_r, AU14\_r, AU15\_r, AU17\_r, AU20\_r, AU23\_r, AU25\_r, AU26\_r, AU45\_r}. 

\subsection{Feature Aggregation over Time Windows}

We now explain the aggregation of feature statistics over time windows with multiple video frames. In this way, the number of features (which was equal to $n$ at the end of the Feature Extraction phase) is increased to $b$ in order to provide more information to the Sequence Model. 

Let the operation of the ``Feature Aggregation over Time Windows'' be shown as
\begin{equation}
    \mathbf{Z}^s_{a \times b} = \textrm{Aggregate}(\mathbf{Y}^s_{m \times n}),
\end{equation}
where $\mathbf{Z}^s_{a \times b}$ is the matrix of the $b$ feature statistics for $a$ aggregated frames. Let $z$ be the number of frames in each time window that are considered for feature aggregation, where $m = a \times z$. Then, in each of $a$ windows, we compute the \textit{mean}, \textit{variance}, \textit{standard deviation}, \textit{minimum}, and \textit{maximum} of each feature over the consecutive $z$ frames resulting in $b$ feature statistics, where $b = 5\times n$. 

\iffalse
\begin{align}
    \label{eq:feature_agg}
    \oplus_{j=0}^a(Min(Y^{ij}_{z \times n}))&=\alpha ^i_{a \times n} \\
    \oplus_{j=0}^a(Max(Y^{ij}_{z \times n}))&=\beta ^i_{a \times n} \\
    \oplus_{j=0}^a(Var(Y^{ij}_{z \times n}))&=\gamma ^i_{a \times n} \\
    \oplus_{j=0}^a(Std(Y^{ij}_{z \times n}))&=\tau ^i_{a \times n} \\
    \oplus_{j=0}^a(Mean(Y^{ij}_{z \times n}))&=\theta ^i_{a \times n} \\
    \alpha ^i_{a \times n} \oplus \beta ^i_{a \times n} \oplus \gamma ^i_{a \times n} \oplus \tau ^i_{a \times n} \oplus \theta ^i_{a \times n}&=Z^i_{a \times b}  \label{eq:feature_agg_end}
\end{align}
\fi

\subsection{Sequence Modeling Combined with Multi-Tasking}

Multi-task learning aims to learn multiple different tasks simultaneously while maximizing performance on one or all of the tasks \cite{caruana1997multitask}. The suggested architecture contains two tasks: The first task is predicting the multi-level engagement label by optimizing the MSE loss between actual and predicted labels. The second task is learning hidden vector embeddings by optimizing the triplet loss.

As shown in Fig.~\ref{fig: triplet_loss}, during sequence modeling, we use three parallel (siamese) Bi-LSTM models with weight sharing to compute the hidden vectors for Triplet Loss and for MSE loss as cascaded to the Fully Connected Neural Network. However, note that training is performed for only one Bi-LSTM model since the Bi-LSTM models in Fig.~\ref{fig: triplet_loss} are used with weight sharing for triplet loss. We call the Bi-LSTM model for the aggregated feature matrix $\mathbf{Z}^s_{a \times b}$ as

\begin{equation}
    T^s_v = \text{Bi-LSTM}(\mathbf{Z}^s_{a \times b}),
\end{equation}
where $T^s_v$ is the hidden vector, which is the hidden state of the last layer of Bi-LSTM model. Thus, the length of this vector, denoted by $v$, is equal to twice the number of hidden units of the last layer of the Bi-LSTM.

\iffalse
\begin{align}
    \label{eq:lstm_formulas}
    i_t &= \sigma(W_{zi}x_t + W_{hi}h_{t-1}) +  W_{ci}c_{t-1} + b_i)\\
    f_t &= \sigma(W_{zf}x_t + W_{hf}h_{t-1}) +  W_{cf}c_{t-1} + b_f)\\
    c_t &= f_t c{t-1} + i_t tanh(W_{zc}x_t + W_{hc}h_{t-1}+ b_c)\\
    o_t &= \sigma(W_{zo}x_t + W_{ho}h_{t-1}) +  W_{co}c_{t-1} + b_o)\\
    h_t &= o_t tanh(c_t) \label{eq:lstm_formulas_end}
\end{align}
\fi

Triplet loss is a loss function where a baseline (anchor) sample is compared with a positive and negative sample. The distance between the anchor and the positive sample is minimized and the distance between the anchor and the negative is maximized. We use the triplet loss function which is presented in \cite{balntas2016learning} and defined as 

\begin{eqnarray}
 &\ell(\textrm{Anchor}, \textrm{Positive}, \textrm{Negative}) = L = \{l_1,\dots l_s \dots,l_S\}^\top,\nonumber\\
 &\quad l_s = \max \{d(\textrm{Anchor}_s, \textrm{Positive}_s) - d(\textrm{Anchor}_s, \textrm{Negative}_s) + {\rm margin}, 0\}, \label{eq:triplet}
\end{eqnarray}
where $S$ is the number of samples in a batch, $d$ is the euclidean distance, and $margin$ is a non-negative margin representing the minimum difference between the positive and negative distances that are required for the loss to be 0. Moreover, $\textrm{Anchor}_s$, $\textrm{Positive}_s$ and $\textrm{Negative}_s$ denote the Anchor, Positive and Negative batches for sample $s$, respectively. 

In addition to the triplet loss, we also minimize the MSE loss which measures the error for the engagement regression. To this end, we cascade the Bi-LSTM model to the Fully Connected Neural Network whose output is the engagement level. Recall that the engagement regression is the main task during the real-time application. Accordingly,  during training, the minimization of MSE can be considered as the main task while the minimization of Triplet loss is the auxiliary task. 
\section{Experimental Results}\label{sec:results}

\subsection{Dataset}
\label{sec:datasets}

\begin{figure}[h!]
\centering{\includegraphics[width=0.8\textwidth, height=4.25cm]{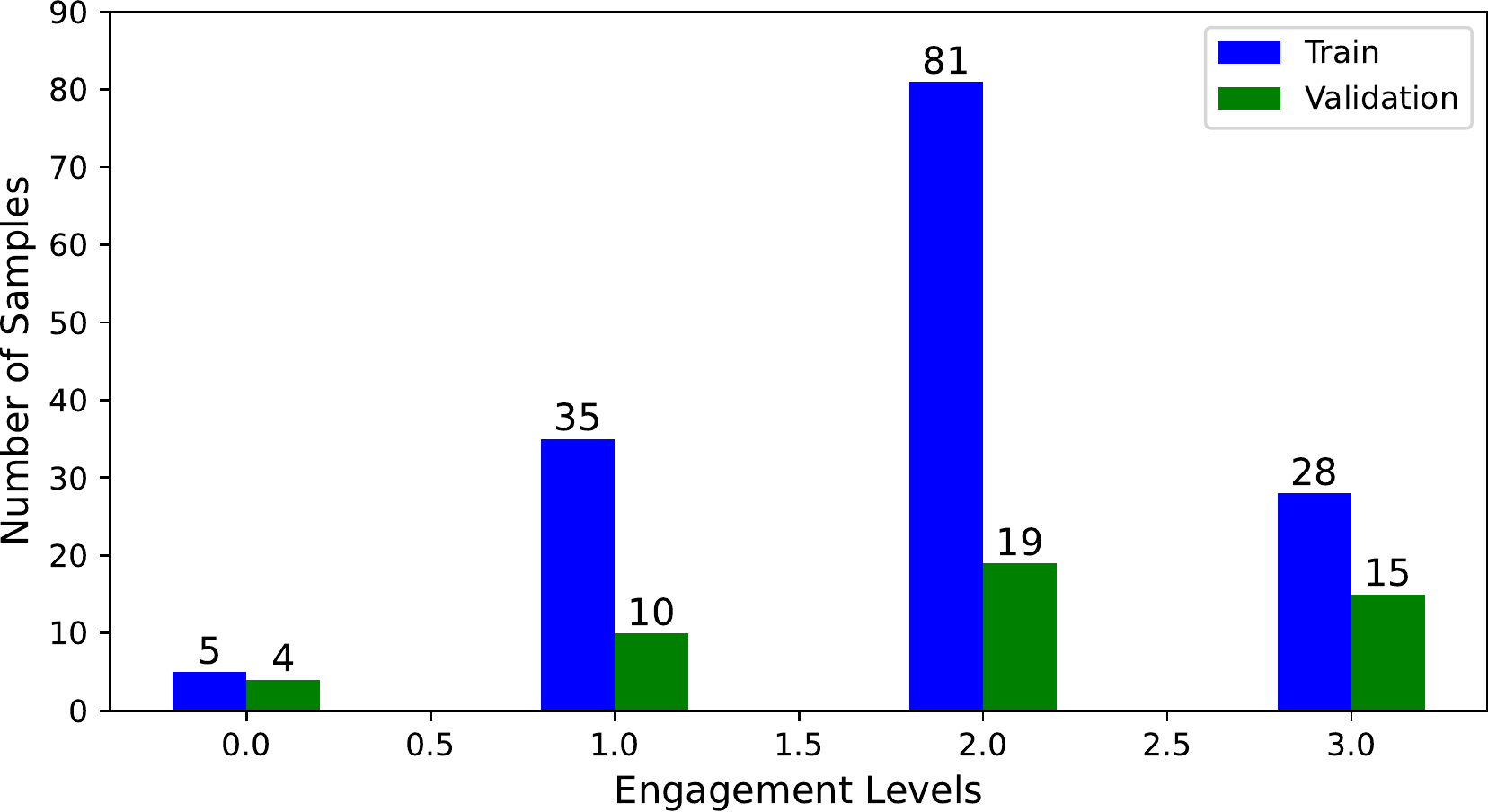}}\vspace{-0.3cm}
\caption{The distribution of the engagement classes for each of the training and validation sets. In this figure, we see that the dataset is highly imbalanced, in particular there is a lack of low level engagement class samples.}
\label{fig:distribution}
\end{figure}

For the performance evaluation of the proposed technique, we use both training and validation datasets published at ``Emotion Recognition in the Wild'' (EmotiW 2020) challenge \cite{dhall2020emotiw} where the engagement regression is a sub-task. The dataset is comprised of 78 subjects (25 females and 53 males) whose ages range from 19 to 27. Each subject is recorded while watching an approximately 5 minutes long stimulus video of a Korean Language lecture. This procedure results in a collection of 195 videos, where the environment varies over videos and the subjects are not disturbed during recording. The engagement level of each video recording is labeled by a team of five between $0$ and $3$ resulting in the distribution shown in Fig.~\ref{fig:distribution}.

\subsection{Experimental Setup and Hyperparameter Settings}

We implemented ED-MTT by using PyTorch on Python 3.7.12. The experiments are executed on the Google Colab platform where the operating system is Linux-5.4.144, and the GPU device is Tesla P100-PCIE-16GB. The model is trained via the adam optimizer \cite{adam} for 500 epochs with $5\times 10^{-5}$ initial learning rate and batch size of $16$.

Furthermore, during our experiments, we first fixed the number of aggregated frames $a = 100$. % with a 20\% percent overlap between aggregated frames. 
At the input of Bi-LSTM, we used a batch normalization with an imbalanced sampler from the ``imbalanced-learn'' library of Python \cite{JMLR:v18:16-365}. Then in order to determine the architectural hyperparameters of the sequential model, we performed a random search for the number of Bi-LSTM layers, the size of the hidden state as well as the number of neurons at each of two fully connected neural network layers. The random search sets are as follows: $\{1, 2, 3\}$ for the number of Bi-LSTM layers, $\{128, 256, 512, 1024\}$ for the size of hidden state of each Bi-LSTM layer\footnote{Note that the size of the hidden state is constant across all Bi-LSTM layers.}, $\{256, 128, 64\}$ for the first layer of the fully connected neural network, and $\{32, 16, 8\}$ for the second layer of fully connected neural network. At the end of this search, the resulting architecture is comprised of 2 Bi-LSTM layers each of whose hidden state size is 1024, and two sequential fully connected layers with $64$ and $32$ neurons respectively.

\subsection{Performance Evaluation}

\begin{table}[h!]
\centering
\caption{\label{emotiw_feature_comp} Performance of The Model Under Different Combinations of Feature Sets}

\begin{tabular}{ |c| c| c| c| c| c|}
\hline
 \textbf{Eye Gaze} & \textbf{Head Pose} & \textbf{Head Rotation} & \textbf{Action Units} & \textbf{MSE} \\ 
 \hline
  \cmark & \xmark  & \xmark & \xmark & 0.08347\\  
 \hline
 \xmark & \cmark  & \xmark & \xmark & 0.07784\\ 
 \hline
 \xmark & \xmark  & \cmark & \xmark & 0.05723\\ 
 \hline
 \xmark & \xmark  & \xmark & \cmark & 0.05044\\ 
 \hline
  \xmark & \xmark  & \cmark & \cmark & 0.06578\\ 
 \hline
 \xmark & \cmark  & \cmark & \xmark & 0.07238\\ 
 \hline
  \cmark & \xmark  & \cmark & \xmark & 0.06915\\ 
 \hline
 \cmark & \cmark  & \xmark & \xmark &  0.06036\\ 
 \hline
 \cmark & \cmark  & \cmark & \xmark & 0.06973\\ 
 \hline
 \cmark & \xmark  & \cmark & \cmark & 0.05681\\ 
 \hline
 \cmark & \cmark  & \xmark & \cmark & \textbf{0.04271}\\ 
 \hline
 \cmark & \cmark  & \cmark & \cmark &   0.05431\\
 \hline
\end{tabular}
\end{table}

We now evaluate the performance of ED-MTT for engagement detection on a publicly available ``Engagement in The Wild'' dataset. During performance evaluation, we first aim to select the subset of facial and head position features with respect to their effects on the performance of our system. To this end, Table~\ref{emotiw_feature_comp} displays the performance of the model under different combinations of feature sets, where the combinations are selected empirically to achieve high performance. Recall that the number of features in each feature set is as follows: 6 features in \emph{Eye Gaze}, 3 features in \emph{Head Pose}, 3 features in \emph{Head Rotation}, and 17 features in \emph{Action Units}. According to our observations on the results presented in this table, we may draw the following conclusions:
\begin{itemize}
    \item The best performance is achieved by using all features except \emph{Head Rotation} features. Accordingly, in the rest of our results, we use the combination of \emph{Eye Gaze}, \emph{Head Pose}, and \emph{Action Unit} features.
    \item The most effective individual feature set is \emph{Action Units}.
    %\item Using \emph{Head Pose} features provides better performance compared with using \emph{Head Rotation}, or \emph{Head Pose} and \emph{Head Rotation} together.
    \item The MSE loss significantly decreases for the majority of the cases when \emph{Action Unit} features are included in the selected features. 
\end{itemize}

\begin{figure}[h!]
\centering{\includegraphics[width=0.8\textwidth]{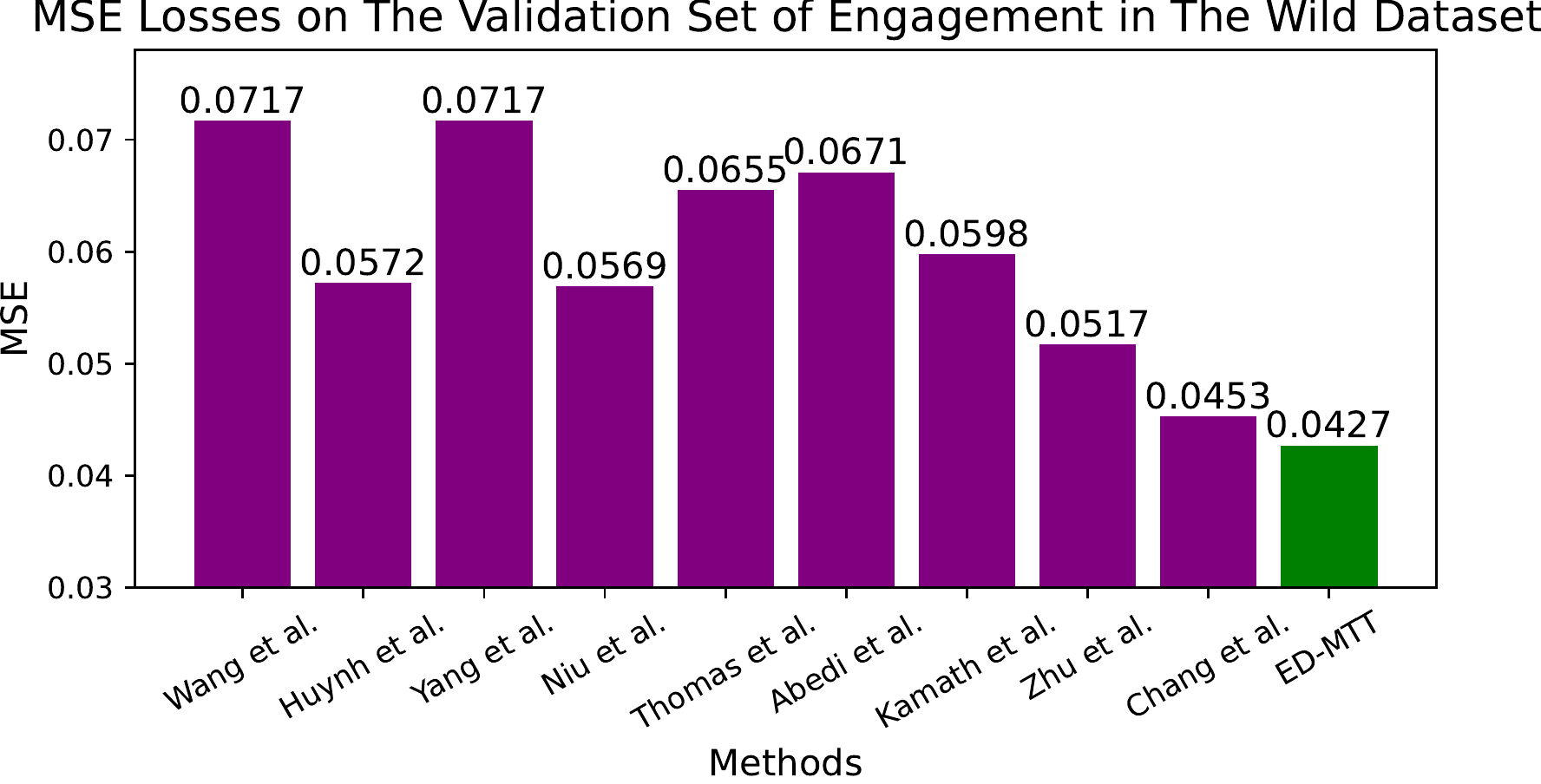}}\vspace{-0.3cm}
\caption{The performance comparison of ED-MTT against the state-of-the-art methods, where the MSE scores are presented as in the original papers. 
%\textit{Wu et al.} \cite{bestemotiw} score: 0.061110,
\textit{Wang et al.} \cite{wang2019bootstrap} score: 0.0717,
\textit{Huynh et al.} \cite{thong2019engagement} score: 0.0572,
%\textit{Wu et al.} \cite{wu2019multi} score: 0.061740,
\textit{Yang et al.} \cite{yang2018deep} score: 0.0717,
\textit{Niu et al.} \cite{niu2018automatic} score: 0.0569,
\textit{Thomas et al.} \cite{thomas2018predicting} score: 0.0655,
\textit{Abedi et al.} \cite{abedi2021affect} score: 0.0671,
\textit{Kamath et al.} \cite{10.1007/978-3-030-82469-3_4} score: 0.0598,
\textit{Zhu et al.} \cite{zhu2020multi} score: 0.0517,
\textit{Chang et al.} \cite{chang2018ensemble} score: 0.0453,
\textbf{\textit{ED-MTT} score: 0.0427}
}
\label{fig:emotiw_paper_scores}
\end{figure}
Furthermore, in Fig.~\ref{fig:emotiw_paper_scores}, we present the comparison of ED-MTT against the state-of-the-art engagement regression methods that are evaluated on the Engagement in The Wild dataset. In this figure, the MSE scores of the state of the art methods are taken from their original papers. The results show that ED-MTT achieves the best performance with $0.0427$ MSE loss on the validation set. Although the performances of all methods are highly competitive with each other, the ED-MTT improved the best performance (Chang et. al. \cite{chang2018ensemble}) in the literature by $6\%$. In addition, the training time of ED-MTT is around 38 minutes for 149 samples for 500 epochs.

\iffalse
Fig.~\ref{fig: train_val_loss} displays the learning curve of our model. Our results in this figure show that ED-MTT does not overfit the training data during the 500 training epochs, which also shows the generalization ability of the resulting ED-MTT is significantly high. 

\begin{figure}[h!]
\centering{\includegraphics[width=\textwidth, height=5.5cm]{Images/emotiw train/mse_losses.png}}
\caption{The validation (blue) and train (orange) losses during the training}
\label{fig: train_val_loss}
\end{figure}
\fi

\begin{figure}[h!]
\centering
\includegraphics[width=\textwidth, height=5cm]{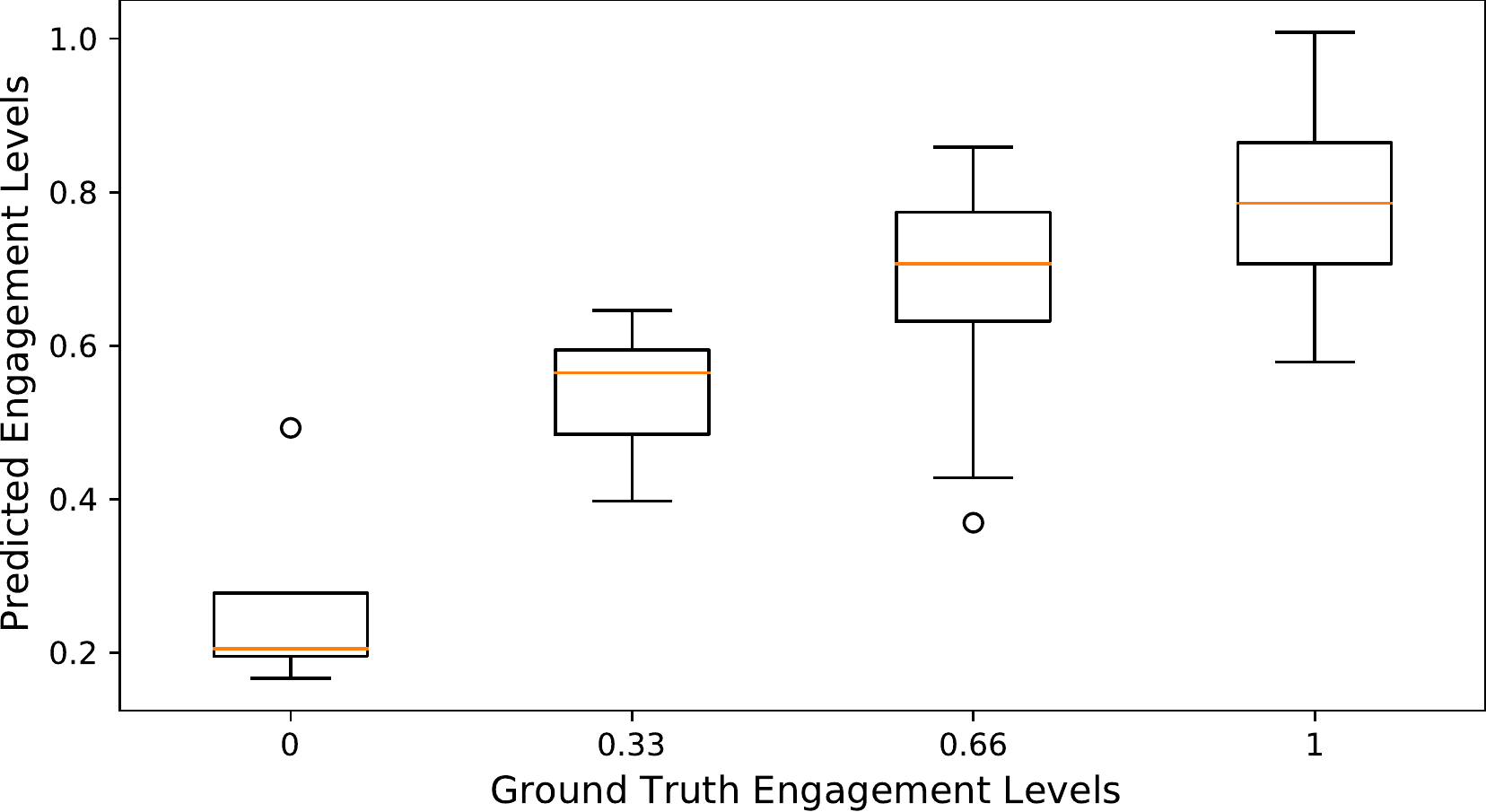}\\\vspace{0.1cm}
\caption{The figure presents (top) the box plot of predicted engagement levels for each class in the ground truth engagement levels.}
\label{fig:boxplot}
\end{figure}

Fig.~\ref{fig:boxplot} displays the box plot of the predicted engagement levels on the validation sets which are classified with respect to the ground truth engagement labels in the dataset. In this figure, from median and percentiles of predicted engagement levels, one may see that the continuous predictions of ED-MTT distinctly reflects the four level of engagement classes in the ground truth labels. Moreover, ED-MTT can easily distinguish between classes 0, 0.33, and 0.66 while the difference between 0.66 and 1.0 is more subtle.

\subsection{Qualitative Results}

\begin{figure}[htbp]
\centering
\includegraphics[width=0.32\textwidth, height=1.8cm]{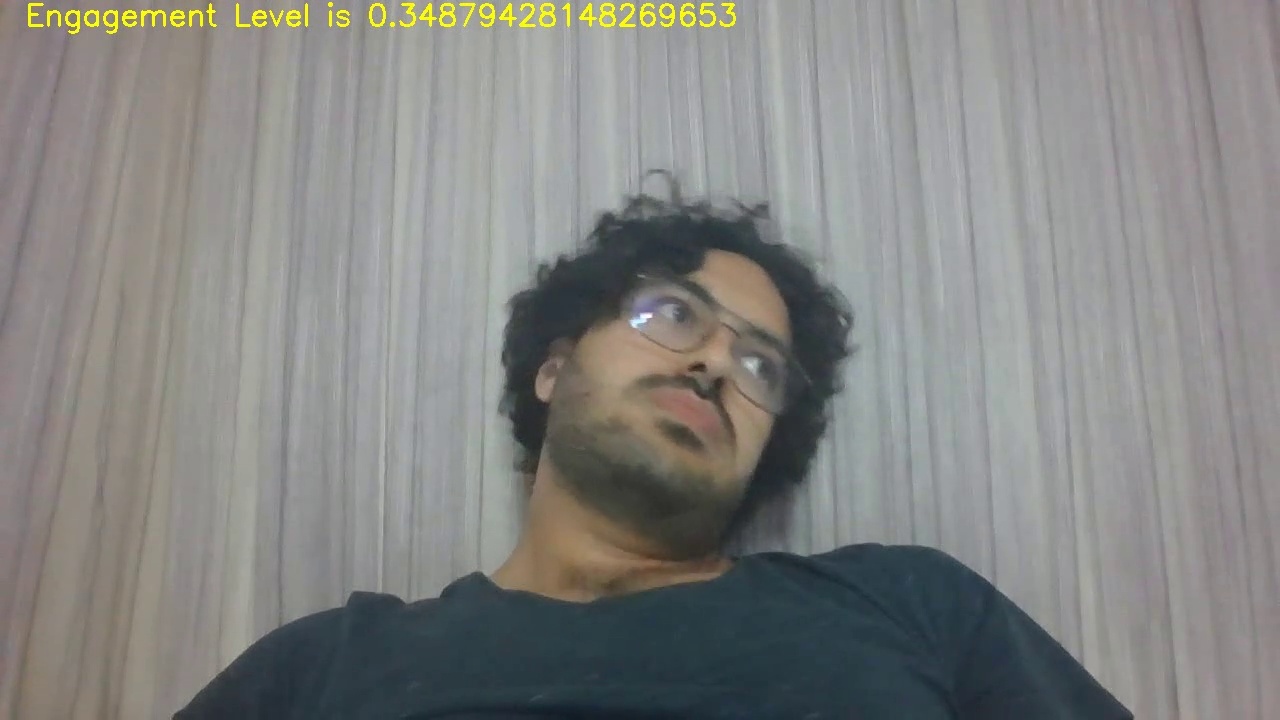}
\includegraphics[width=0.32\textwidth, height=1.8cm]{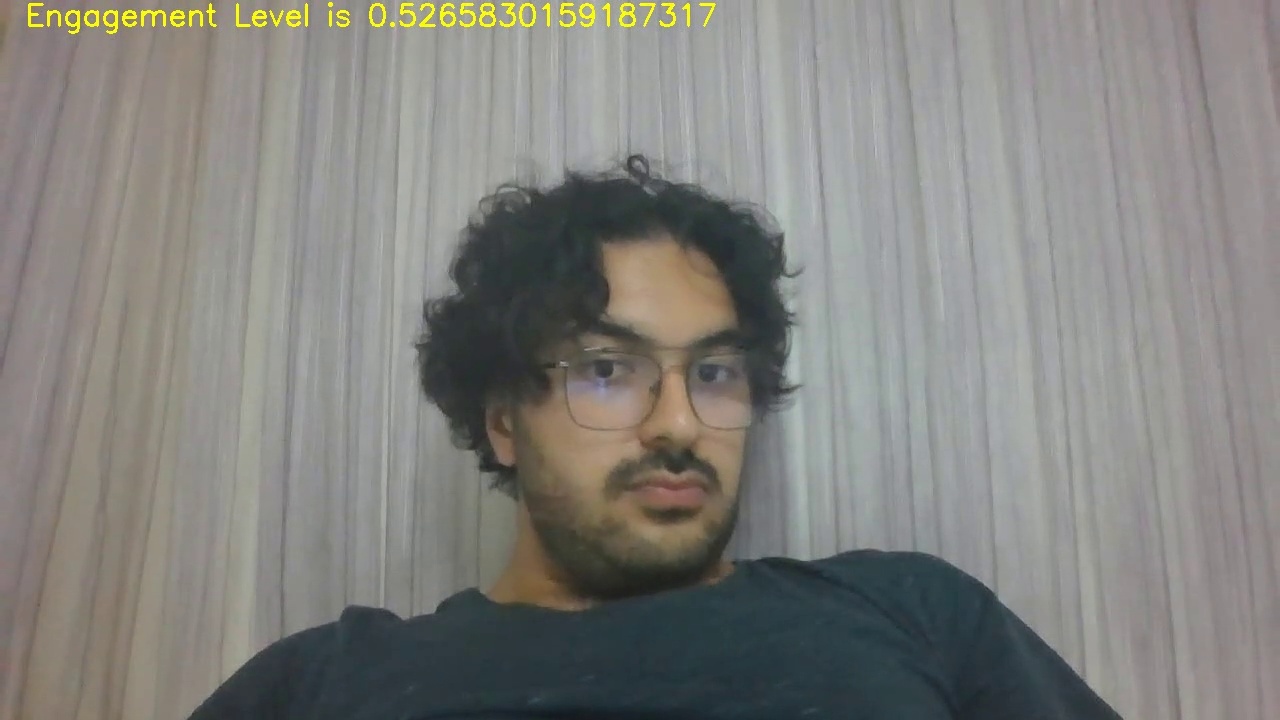}
\includegraphics[width=0.32\textwidth, height=1.8cm]{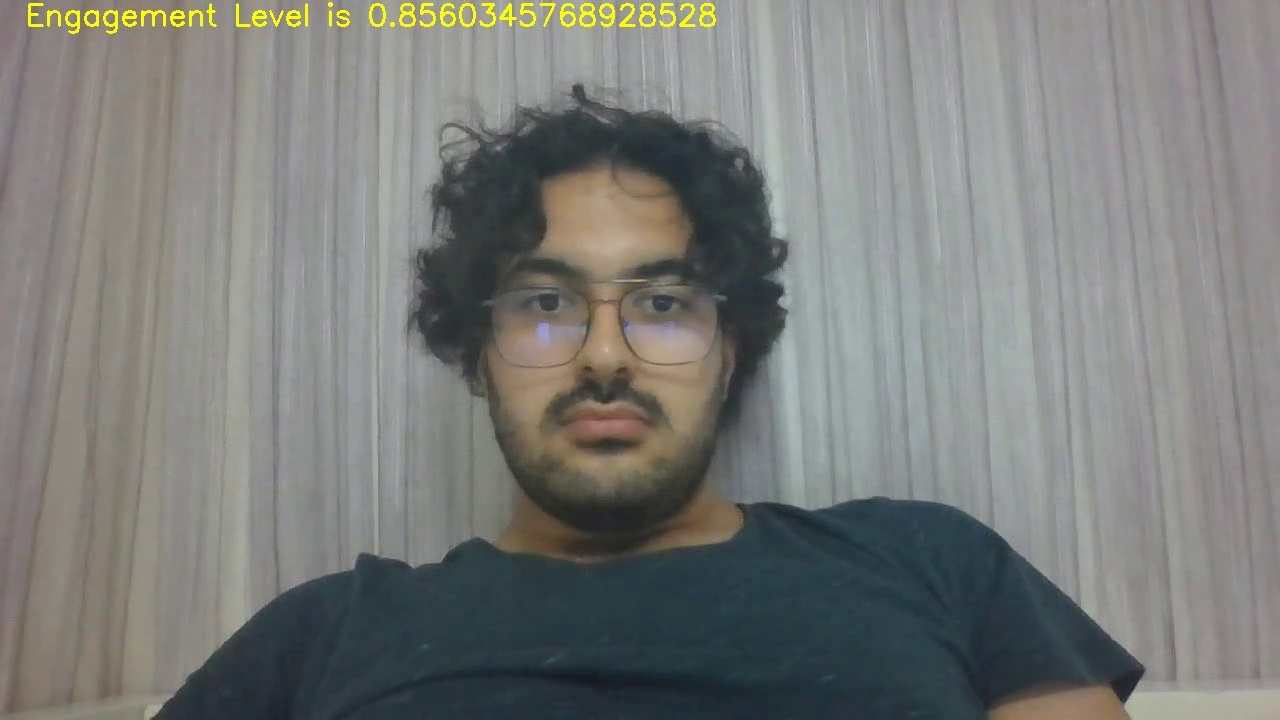}\\

\includegraphics[width=0.32\textwidth, height=1.8cm]{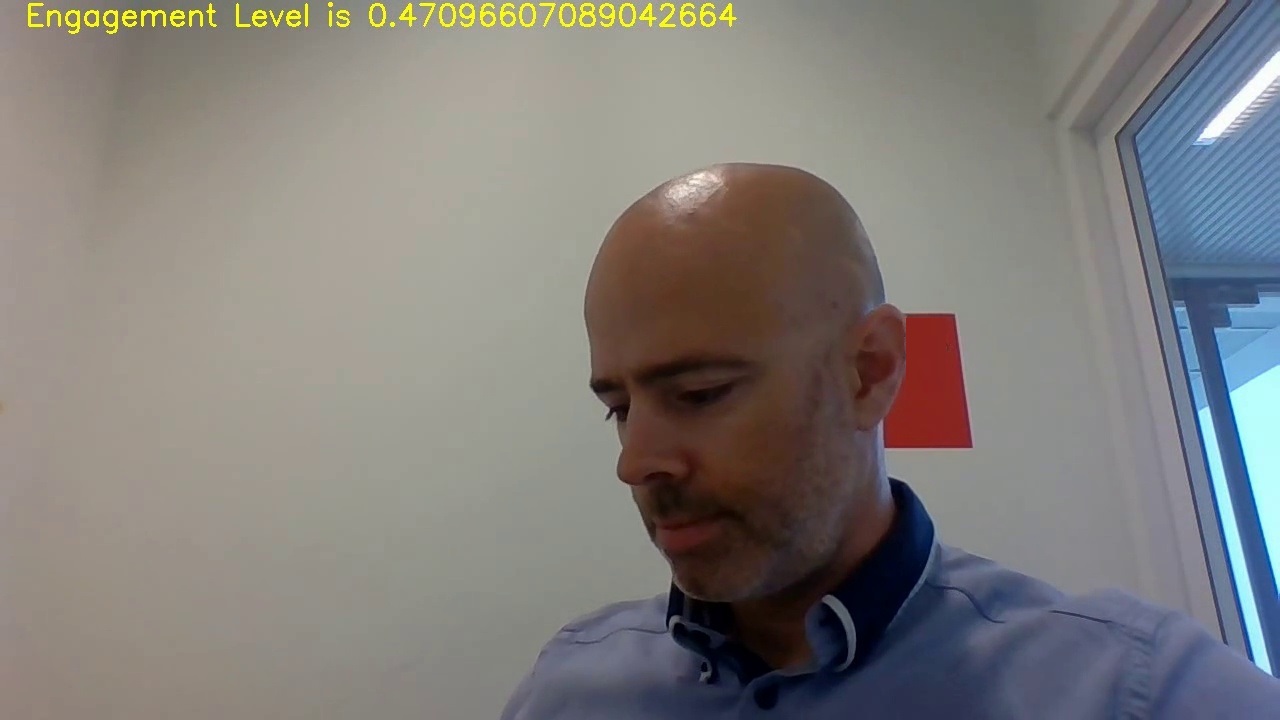}
\includegraphics[width=0.32\textwidth, height=1.8cm]{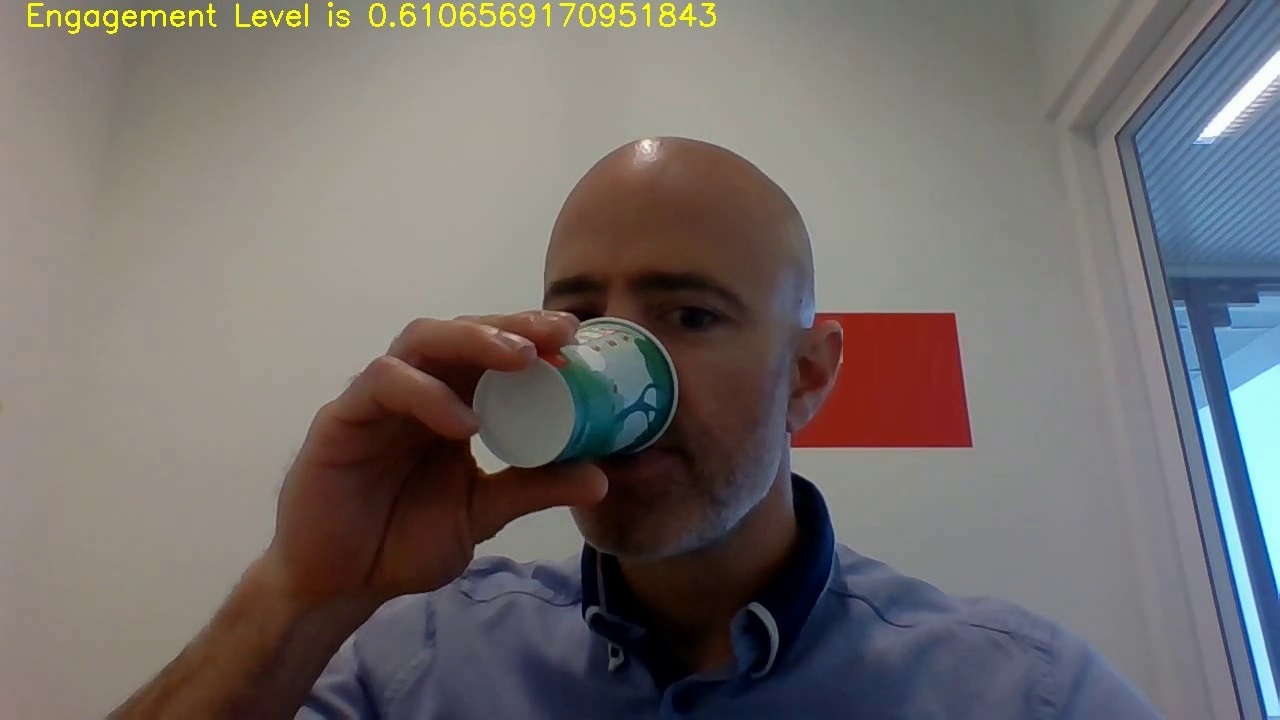}
\includegraphics[width=0.32\textwidth, height=1.8cm]{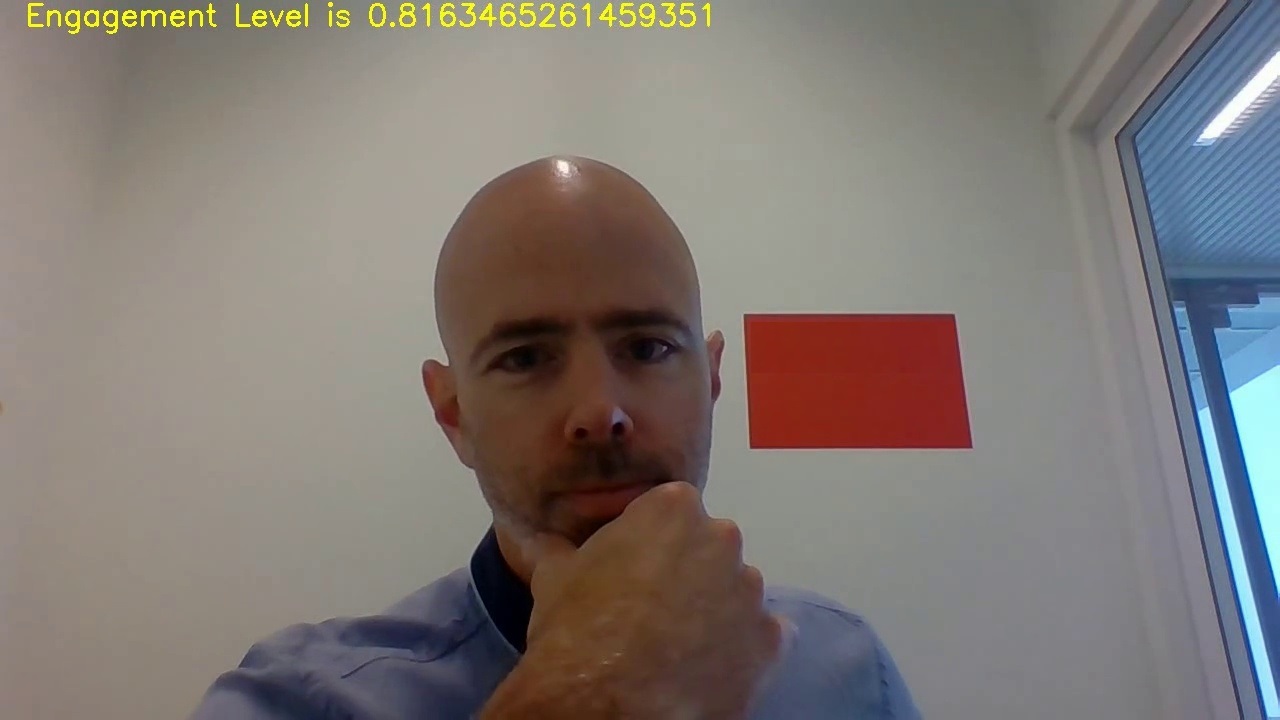}
\caption{Sample images with the following predicted engagement levels by ED-MTT: 0.35 (top left), 0.53 (top middle), 0.86 (top right), 0.47 (bottom left), 0.61 (bottom middle), and 0.82 (bottom right).}
\label{fig:real_life}
\end{figure}

Finally, ED-MTT is also tested on a preliminary real-life engagement detection tasks for which the prediction results are presented in Fig.~\ref{fig:real_life}. These results show that the proposed model, ED-MTT, trained on Engagement in The Wild dataset is able to provide highly successful predictions in real-life use-cases, which are totally different than the cases in the training set. According to our observations on the prediction results for a total (approximately) 12 minutes long videos including 8 people, the model can successfully distinguish different levels of engagement (very low, low, high, and very high engagement levels). However, the predicted engagement levels lie between $0.2$ and $0.92$, which forces to determine smaller quantization intervals to classify engagement levels in real-life use-cases.

\section{Conclusion}\label{sec:conclusion}

Online working and learning environments are currently more essential in our lives, especially after the COVID-19 era. In order to improve the user experience and efficiency, advanced tools, such as recognition of user interaction, became highly crucial in these digital environments. For e-learning, one of the most important tools might be the engagement detection system since it provides valuable feedback to the instructors and/or students.  

In this paper, we developed a novel engagement detection system called ``ED-MTT'' based on multi-task training with triplet and MSE losses. For engagement regression task, ED-MTT uses the combination of \emph{Eye Gaze}, \emph{Head Pose}, and \emph{Action Units} feature sets and is trained to minimize MSE and triplet loss together. This training approach is able to improve the regression performance due to the following reasons; 1) multi-task training with two losses introduces an additional regularization and reduces over-fitting due to very small sample size, 2) triplet loss measures relative similarity between samples to mitigate the label reliability problem. 3) minimization of MSE ensures that the main loss considered for the regression problem is minimized alongside the triplet loss. 

The performance of ED-MTT is evaluated and compared against the performances of the state-of-the-art methods on the publicly available Engagement in The Wild dataset which is comprised of separated training and validation sets. Our results showed that the novel ED-MTT method achieves $6\%$ lower MSE than the lowest MSE achieved by the state-of-the-art while the training of ED-MTT takes around $38$ minutes for $149$ samples for $500$ epochs. We tested the performance of ED-MTT for real-life use cases with 8 different participants, and the prediction results for majority of these cases were shown to be highly successful.  

%Future works may focus on the automation of the feature aggregation and selection operations combined with the sequence model.

\bibliographystyle{splncs04}
\bibliography{references}
\end{document}